\documentclass[11pt]{article}

\usepackage[preprint]{acl}
\usepackage{times}
\usepackage{latexsym}
\usepackage[T1]{fontenc}
\usepackage[utf8]{inputenc}
\usepackage{microtype}
\usepackage{inconsolata}
\usepackage{graphicx}
\usepackage{amsmath}
\usepackage{amssymb}
\usepackage{booktabs}
\usepackage{multirow}
\usepackage{xcolor}
\usepackage{algorithm}
\usepackage{algorithmic}
\usepackage{subcaption}
\usepackage{makecell}
\usepackage{colortbl}
\usepackage{tcolorbox}
\tcbuselibrary{skins, breakable}
\usepackage{enumitem}
\usepackage{placeins}
\usepackage{float}
\usepackage{caption}
\usepackage{placeins}

\newcommand{\spm}[1]{\ensuremath{{\scriptstyle \pm #1}}}

\title{Attention-Aware Routing: Coupling Routing and Attention in MoEs}

\newcommand\blfootnote[1]{%
  \begingroup
  \renewcommand\thefootnote{}\footnote{#1}%
  \addtocounter{footnote}{-1}%
  \endgroup
}

\author{
  \textbf{Despoina Kosmopoulou}$^{1}$ \quad
  \textbf{Anastasios Tsetsilas}$^{1}$ \quad
  \textbf{Efthymios Georgiou}$^{2,*}$ \\
  \textbf{Giannis Karamanolakis}$^{3,\dagger}$ \quad
  \textbf{Swastik Roy}$^{3,\dagger}$ \quad
  \textbf{Alexandros Potamianos}$^{1,4}$ \\
  $^{1}$National Technical University of Athens \\
  $^{2}$University of Bern \\
  $^{3}$Amazon AGI \\
  $^{4}$Archimedes RU, Athena RC \\
  \texttt{despoinakkosmopoulou@gmail.com}
}
\begin{document}
\maketitle
\blfootnote{$^{*}$Work done while at NTUA.}
\blfootnote{$^{\dagger}$This work is independent of and outside of the work at Amazon.}

\begin{abstract}
In Mixture-of-Experts language models, the router typically selects and weights experts  based on the token's hidden state,
utilizing limited contextual information. We propose \textbf{Attention-Aware Routing (AAR)}, which augments the router with temporal and spectral features extracted from a sliding window of attention weights that represent a summary of the model's \textit{contextual} state, disentangled from the hidden state.
Keeping the base transformer entirely frozen, we train only the routing parameters, isolating routing as the sole variable. AAR improves GSM8K by $+3.37$ pp over a routing-only SFT baseline on OLMoE.
Beyond performance, we show that routing and attention form a coupled circuit: routing changes at layer $l$ propagate through the residual stream to amplify attention sinks at layer $l+1$, 
reshaping attention without any direct update to the attention mechanism itself. Further, AAR reduces long diverging generation, with incorrect answers getting shorter, while correct answers remain unchanged in length.
Finally, AAR is strongly depth-sensitive: applying it indiscriminately across layers can degrade factual retrieval, whereas mathematical reasoning gains persist when it is introduced deeper in the network. This sensitivity exposes a retrieval--reasoning tension across depth and makes layer-selective AAR a controlled probe of the routing-relevant information carried by attention at different layers.
\end{abstract}


\section{Introduction}
\label{sec:intro}

Mixture-of-Experts (MoE) architectures \citep{shazeer2017outrageously, fedus2022switch, lepikhin2021gshard} scale language model capacity while keeping inference cost manageable by activating only a subset of experts per token. The router, the mechanism that decides which experts process each token, is therefore a crucial component \citep{dikkala-etal-2023-benefits}: it not only ensures the efficient allocation of computation, but determines which computational pathways shape each token's representation at every layer.

Standard routers make this decision typically utilizing the hidden state as their sole input. Given a token's hidden state $\mathbf{h}_i^l$, the router computes a linear projection and selects the top-$k$ experts to process it \citep{shazeer2017outrageously}.
The hidden state carries contextual information, entangled with other content information within the hidden state. 
In this work, we investigate how a model's performance and behavior can change, if disentanglement of content and contextual relational information is enforced as an inductive routing bias. 

To achieve a simple form of contextual disentanglement, the router can be augmented with a window of recent attention weights. While hidden states summarize what each token represents, attention weights encode compressed relational intra-sequence structure: which positions the model attends to, which it ignores, and where focus concentrates.

We propose \textbf{Attention-Aware Routing (AAR)}, which augments the standard router with a sliding window of attention weights. The attention weights provide a short summary of how attention was distributed in the latest generation steps. Training only the routing parameters while keeping the base transformer frozen, AAR isolates routing as the sole variable. Our main contributions are:


\begin{enumerate}
    \item \textbf{AAR improves mathematical reasoning.} We show that endowing the router with contextual information directly improves downstream performance. AAR improves GSM8K by $+3.37$ pp over routing-only SFT on OLMoE \citep{muennighoff2024olmoe}, with consistent gains on mathematical reasoning across benchmarks, \textit{achieved by training the routers alone}.
    
    \item \textbf{AAR enables and exploits a routing-attention coupling.} AAR modifies generation from routing alone.
    Internally, routing changes at layer $l$ propagate through the residual stream to amplify attention sinks at layer $l+1$, showing that routing and attention form a coupled circuit.
    Externally, AAR reduces diverging generation length: incorrect answers get shorter while correct answers are unaffected.

\item \textbf{AAR reveals a depth-dependent retrieval--reasoning
trade-off.} AAR is not equally beneficial across layers: early-layer
interventions can disrupt factual retrieval, while mathematical
reasoning gains persist when AAR is applied deeper in the network.
Across models, middle-to-deep layers provide a useful starting point
for effective AAR placement. This depth sensitivity also turns
layer-selective AAR into a controlled probe of where routing-relevant
information is encoded in attention across the network.

\end{enumerate}
\section{Related Work}
\label{sec:related}

\paragraph{MoE routing.}
Standard token-choice routing computes a softmax over expert logits derived from the token representation and selects the top-$k$ experts \citep{shazeer2017outrageously}.
Load-balancing auxiliary losses encourage uniform expert utilization \citep{fedus2022switch}.
Expert-choice routing \citep{zhou2022mixtureofexpertsexpertchoicerouting} inverts the assignment.
Our work keeps token-choice routing intact and augments it with an attention-conditioned auxiliary signal.

\paragraph{Various Routers.}
\citet{qiu2025layerwise} implement MoE routing with a recurrent unit, shared across layers. The input to the router at each layer comes from the hidden state at the current layer, as well as previous routing decisions (the recurrent unit's hidden state). In \citet{do-etal-2023-hyperrouter} a fixed randomly initialized hypernetwork generates the router's parameters, conditioned on a trainable router embedding. The HyperRouter alleviates representation collapse, achieving low-entropy routing, and limits the number of experts used during inference. Similar to these works, our routing method does not rely on the hidden state alone to generate each layer's routing logits, but incorporates attention weights as an additional signal.

\paragraph{Attention sinks.}
\citet{xiao2024efficient} observe that transformers concentrate
disproportionate attention mass on the first token (the ``attention sink''),
which stabilizes representations under streaming contexts.
\citet{barbero2025why} show that sink presence slows information mixing and increases robustness to prompt perturbations.
We find that AAR systematically strengthens attention sinks in layers following the routing modification, linking this to improved reasoning.

\paragraph{Layer-wise functional specialization.}
\citet{decoupling}, \citet{song2026demystifyingrolesllmlayers}, show that shallow LLM layers dominate likelihood and retrieval tasks while mid-to-deep layers are essential for reasoning and generation.
\citet{yang-etal-2025-unveiling} distinguish latent reasoning from factual shortcuts across depth.

\paragraph{Attention-aware interventions for reasoning.}
\citet{nguyen-etal-2026-improving} improve chain-of-thought reasoning through attention-aware interventions at inference time, based on empirical hard-written rules.
Our work differs in that the attention signal is incorporated into the routing mechanism during training, making the effect persistent, learnable and architecture-native.

\paragraph{Overthinking in chain-of-thought.}
\citet{li2026sample}, \citet{su2025underthinkingoverthinkingempiricalstudy} show that longer reasoning chains correlate with higher error rates.
Our finding that AAR reduces output length specifically on incorrect answers connects routing-induced changes to this line of research.
\section{Method}
\label{sec:method}
\subsection{Standard MoE Routing}
In a typical MoE layer of a transformer LLM at depth $l$, given a token representation $\mathbf{h}_i^l$ after the attention sub-layer, the router computes expert selection logits as $\mathbf{g}_i = W_r \mathbf{h}_i^l$, where $W_r \in \mathbb{R}^{E \times d}$ and $E$ is the number of experts. Top-$k$ experts are selected after applying a softmax over $\mathbf{g}_{i}$, and the token is processed by a weighted combination of the selected experts' outputs \citep{shazeer2017outrageously}.

\subsection{Attention-Aware Routing}
Attention-Aware Routing operates in parallel to the standard router.
In each selected MoE layer $l$, an untrained routing module is introduced, $A_l$. Instead of the hidden state, this module routes based on a short window of the token's attention weights to previous tokens of the sequence.

\begin{figure}[t]
    \centering
    \includegraphics[width=1.1\columnwidth]{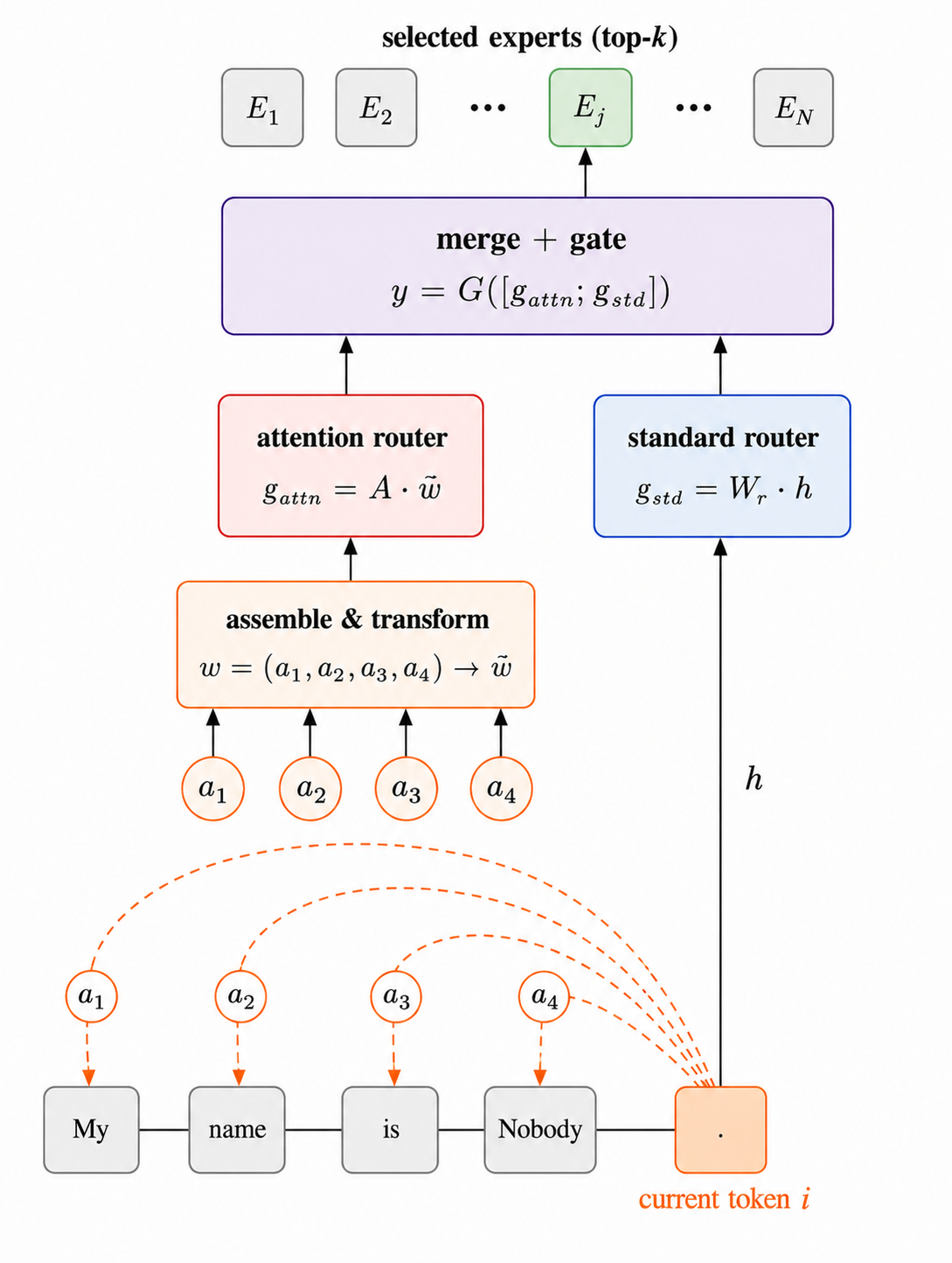}
    \caption{\textbf{Attention-Aware Routing.} For the current token $i$, attention weights to the previous $M$ tokens are averaged across heads and assembled into a sliding window $\mathbf{w}$, then transformed into the augmented window $\tilde{\mathbf{w}} = \phi(\mathbf{w})$. The attention router $A_l$ maps $\tilde{\mathbf{w}}$ to expert logits $\mathbf{g}^{\text{attn}}$, which are merged and gated with the standard router's logits $\mathbf{g}^{\text{std}} = W_r \mathbf{h}$ to produce the final routing decision over experts $E_1, \ldots, E_N$.}
    \label{fig:aar_architecture}
\end{figure}

\paragraph{Attention window.}
The attention window is a sliding window of size $M$, including only attention weights to the last $M$ tokens in the token's context. For the cases there are less than $M$ tokens early in the sequence, the window is zero-padded from the left, keeping its length constant.
The attention weights to the tokens in the last $M$ positions are averaged across the attention heads, assembling into a 1-dimensional window consisting of one attention weight per previous token. We denote the resulting raw window by $\mathbf{w}_i^l \in \mathbb{R}^{M}$.

\paragraph{Augmented attention window.}
The attention weights can be further transformed post assembly, to allow learning of more features by the linear routing layer. Specifically, the augmented attention window $\tilde{\mathbf{w}}_i^l$ is obtained by applying a fixed, non-learned feature map $\phi$ to the raw window,
\begin{equation}
    \tilde{\mathbf{w}}_i^l = \phi(\mathbf{w}_i^l) \in \mathbb{R}^{M'}.
\end{equation}
The map $\phi$ collects one or more views of the window. In the time domain, $\phi$ retains the raw window as-is (AAR$^\text{T}$). We additionally consider a frequency-domain view: the DFT magnitude of the window, which is shift-invariant and exposes spectral structure that a single linear layer could not otherwise recover from the raw window. Concatenating both views,
\begin{equation}
    \phi(\mathbf{w}_i^l) = \big[\, \mathbf{w}_i^l \;\|\; \log\!\big(1 + |\mathcal{F}(\mathbf{w}_i^l)|\big) \,\big],
\end{equation}
where $\mathcal{F}$ denotes the DFT, we retain the first $K = \lfloor M/2 \rfloor + 1$ magnitude bins. The $\log(1+\cdot)$ compresses the dynamic range of the magnitudes before they enter the router. This gives $M' = M + K$; the time-only variant is recovered by setting $\phi(\mathbf{w}_i^l) = \mathbf{w}_i^l$, for which $M' = M$.

\paragraph{Attention router.}
The augmented window is mapped to expert logits by a single linear router,
\begin{equation}
    \mathbf{g}^{\text{attn}}_i = A_l^\top \tilde{\mathbf{w}}_i^l, \qquad A_l \in \mathbb{R}^{M' \times E}.
\end{equation}
We take $A_l$ to be block-diagonal across the views assembled by $\phi$, with one block per view, so that each view is routed by its own block and no cross-view weights are introduced.

\paragraph{Router interpolation.}
Each view is gated by a confidence weight conditioned on the hidden state and that view's window. With $\mathbf{g}^{\text{attn}}_i \in \mathbb{R}^{2E}$ the per-view attention logits and $G_l$ block-diagonal,
\begin{equation}
    \boldsymbol{\alpha}_i = \sigma\!\big(G_l\,[\, \mathbf{h}_i^l \;\|\; \tilde{\mathbf{w}}_i^l \,]\big) \in \mathbb{R}^{2E},
\end{equation}
so each gate sees only the hidden state and its own window. Setting $B = \tfrac{1}{2}[\, I_E \;\|\; I_E \,]$ and $\boldsymbol{\beta}_0 = \mathbf{1} - B\boldsymbol{\alpha}_i$, the standard router keeps the remaining per-expert weight and the final logits are
\begin{equation}
    \mathbf{g}_i = \boldsymbol{\beta}_0 \odot \mathbf{g}^{\text{std}}_i + B\,\big(\boldsymbol{\alpha}_i \odot \mathbf{g}^{\text{attn}}_i\big).
\end{equation}
Standard routing is recovered when $\boldsymbol{\alpha}_i \equiv \mathbf{0}$.

\section{Experiments}
\label{sec:experiments}
\begin{table*}[t]
\centering
\resizebox{\textwidth}{!}{%
\begin{tabular}{lcccccc}
\toprule
\multirow{2}{*}{Configuration} & \multicolumn{1}{c}{GSM8K} & \multicolumn{1}{c}{BBH} & \multicolumn{1}{c}{MMLU} & \multicolumn{1}{c}{MMLU-Math} & \multicolumn{1}{c}{HumanEval} & \multicolumn{1}{c}{IFEval} \\
 & 8-shot CoT acc. & 0-shot acc. & 5-shot acc. & 5-shot acc. & pass@10 & loose acc. \\
\midrule
OLMoE-0125-SFT         & 51.33 & \textbf{36.80} & 53.96 & 31.01 & 41.46 & 56.19 \\
+ routing-only SFT     & 51.99\spm{0.76} & 36.17\spm{0.45} & 53.91\spm{0.01} & 30.45\spm{0.56} & \textbf{41.46}\spm{0.80} & 56.56\spm{1.09} \\
+ AAR$_S$, $M=40$   & \textbf{55.36}\spm{0.74} & 36.71\spm{0.46} & \textbf{53.96}\spm{0.20} & 32.91\spm{0.54} & 40.00\spm{1.26} & 55.64\spm{0.41} \\
+ AAR$_S$, $M=20$   & 54.89\spm{1.02} & 36.30\spm{0.42} & 53.93\spm{0.10} & \textbf{33.48}\spm{0.49} & 40.98\spm{1.80} & \textbf{56.67}\spm{1.36} \\
+ AAR$^{\text{T}}_S$, $M=20$    & 54.57\spm{1.10} & 36.33\spm{0.61} & 53.78\spm{0.14} & 32.29\spm{0.64} & 40.85\spm{1.14} & 56.34\spm{1.88} \\
\midrule
Qwen1.5-MoE-A2.7B-Chat & & & & & & \\
+ routing-only SFT     & 53.48\spm{1.45} & \textbf{40.96}\spm{0.43} & \textbf{59.22}\spm{0.04} & \textbf{42.06}\spm{0.43} & 39.26\spm{0.83} & \textbf{36.45}\spm{1.26} \\
+ AAR$_{S_{Q}}$, $M=40$    & \textbf{55.36}\spm{1.41} & 40.51\spm{0.36} & 59.02\spm{0.13} & {41.40}\spm{0.41} & \textbf{40.48}\spm{2.27} & 35.82\spm{0.97} \\
\bottomrule
\end{tabular}}
\caption{AAR compared against trained baselines. GSM8K consistently improves, with full AAR achieving the best results. $S = \{9, \ldots, 15\}$ for OLMoE. 
$S_Q = \{14, \ldots, 17\}$ for Qwen1.5. Best in bold.
}
\label{tab:main}
\end{table*}

\subsection{Setup}
\paragraph{Model.} For the main experiments, we use OLMoE-1B-7B-SFT \citep{muennighoff2024olmoe} \footnote{\url{allenai/OLMoE-1B-7B-0125-SFT}}, a 7B-parameter MoE model with 1B active parameters per token, 16 MoE layers, and 64 experts per layer (top-$k$ routing with $k=8$).

\paragraph{Training.} We fine-tune only the routing parameters: the original router weights $W_r^l$ and the auxiliary parameters $(A_l, G_l)$, while keeping all other model parameters frozen. We use the AdamW optimizer \citep{loshchilov2018decoupled} with a batch size of $1024$, sequence length of $512$, a linear learning rate schedule with LR $= 1 \times 10^{-4}$, and warmup ratio $0.03$. Each configuration is trained with 5 random seeds. Training is performed on the Leonardo supercomputer at CINECA, using 4$\times$A100 64GB GPUs; a single OLMoE run takes approximately 4 GPU hours.

\paragraph{Data.} We fine-tune on a 25\% sample ($\sim$220K examples) of Tulu3 \citep{lambert2025tulu}, the SFT dataset used for OLMoE, including high-quality math, code, and instruction-following samples.

\paragraph{Baselines.} We compare AAR against the original OLMoE-0125-SFT model and identical routing-only SFT runs without the auxiliary routers, isolating the effect of the additional attention signal from that of continued training. We verify our results with the same comparisons on Qwen1.5 \citep{qwen_moe}\footnote{\url{Qwen/Qwen1.5-MoE-A2.7B-Chat}} (24 layers, 2.7B active parameters, 14.3B in total), and mathematics-focused evaluation for Qwen3.6 \citep{qwen36_35b_a3b} (40 layers, 3B active parameters, 35B in total)
with their corresponding chat templates.

\paragraph{Evaluation.} We evaluate on GSM8K \citep{cobbe2021gsm8k}, BBH \citep{suzgun2022bbh}, MMLU \citep{hendrycks2021measuring}, MATH-500 \citep{lightman2023lets}, HumanEval \citep{chen2021humaneval}, and IFEval \citep{zhou2023ifeval}.

\subsection{Main Results}

Table~\ref{tab:main} consolidates results across reasoning and factual benchmarks. AAR, applied to layers $S = \{9, \ldots, 15\}$ with $M=40$, achieves the best GSM8K performance at $55.36\pm0.74$, a gain of $+3.37$ pp over routing-only SFT. AAR$^{\text{T}}$ (AAR utilizing only time-domain weights) at $M=20$ already improves GSM8K from $51.99$ to $54.57$ ($+2.58$ pp), with the full AAR providing further gain. Performance on BBH, HumanEval, and IFEval is maintained within noise for both. GSM8K gains replicate on Qwen1.5 and are reported in Table~\ref{tab:main}.

\paragraph{Generalization across benchmarks and scale.}
We next test how AAR's mathematical reasoning gains extend beyond GSM8K and persist at a larger model scale. On MATH-500, AAR improves OLMoE by $+2.64$ pp over routing-only SFT (Table~\ref{tab:math500}), showing that the gains extend to a distinct mathematical reasoning benchmark.

We further evaluate AAR on Qwen3.6-MoE, a larger 3B-active/35B-total model with 40 layers. We apply AAR ($M=40$) to full-attention layers $\{23,27,31\}$, located in the middle-to-deep portion of the network, and train on 12.8K samples without architecture-specific tuning. AAR improves MATH-500 by $+1.13$ pp over three seeds (Table~\ref{tab:math500}), despite the substantially stronger baseline and reduced training budget. Together, these results show that AAR's mathematical reasoning gains are not specific to GSM8K and persist on a larger model.

\begin{table}[t]
\centering
\small
\begin{tabular}{lc}
\toprule
Method & MATH-500 \\
\midrule
\multicolumn{2}{l}{\textit{OLMoE}} \\
Routing-only SFT & 14.00 $\pm$ 1.83 \\
AAR ($L9$--$15$) & \textbf{16.64} $\pm$ 0.91 \\
\midrule
\multicolumn{2}{l}{\textit{Qwen3.6-MoE}} \\
Routing-only SFT & 78.47 $\pm$ 1.30 \\
AAR ($L23,27,31$) & \textbf{79.60} $\pm$ 1.51 \\
\bottomrule
\end{tabular}
\caption{MATH-500 accuracy for AAR$^{\text{TF}}$ and routing-only
SFT. OLMoE results are averaged over 5 seeds and Qwen3.6-MoE
results over 3 seeds.}
\label{tab:math500}
\end{table}

\subsection{Window Size and Signal Structure}
\label{sec:window}

\paragraph{Window size.}
We find that medium-sized windows, $M = 20$--$40$, work best for Attention-Aware Routing. Table~\ref{tab:window} reports GSM8K accuracy as a function of $M$ for AAR$^{\text{T}}$, with $S = \{9, \ldots, 15\}$: AAR$^{\text{T}}$ peaks at $M=20$, and degrades for large windows. AAR reaches its best performance at $M=40$ (Table~\ref{tab:main}).

\begin{figure}[!hp]
    \centering
    \includegraphics[width=\columnwidth]{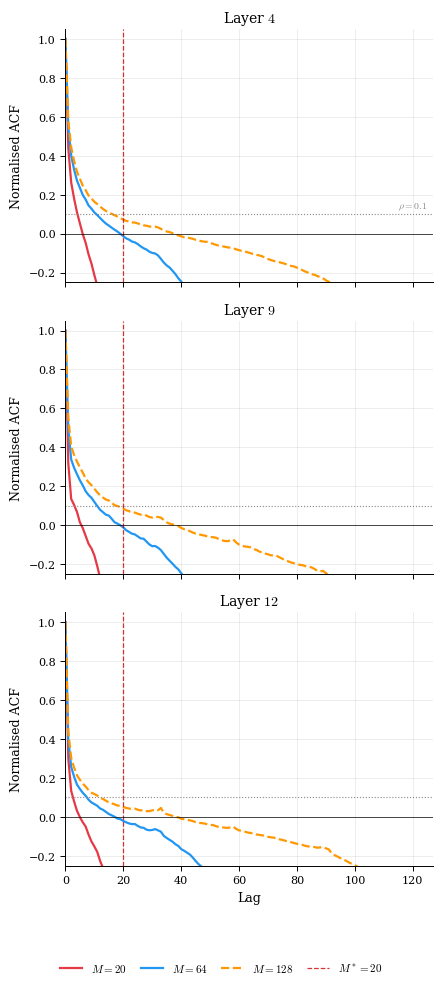}
    \caption{Normalized autocorrelation function (ACF) of the attention window signal $w^l_i(t)$ at layers 4, 9, and 12, for example window sizes $\in \{20, 64, 128\}$. The ACF is averaged over query tokens and prompts from GSM8K (5-shot). The dashed vertical line marks lag $= 20$. Across all layers and window sizes, the ACF decays below $0.1$ by lag $\sim\!20$, revealing an intrinsic correlation length of $20$ tokens in the attention signal. A window of $M=20$ is thus sufficient to capture most structured information.}
    \label{fig:acf_curves}
\end{figure}

\paragraph{Intrinsic correlation length of the attention signal.}
The optimal window size for AAR$^{\text{T}}$ is explained by the temporal structure of the attention signal itself. We analyze $\mathbf{w}_i^l$ via its normalized autocorrelation function (ACF), computed for GSM8K prompts and averaged over query tokens. In Figure~\ref{fig:acf_curves} we plot the ACF and mark the lag at which it first decays below $0.1$, across layers and window sizes $M$. Even at $M=128$, the \textit{effective memory length} saturates at roughly 20 tokens across layers: beyond this lag, successive attention weights are uncorrelated. This reveals an intrinsic correlation length of $\sim$20 tokens in the attention signal, so for AAR$^{\text{T}}$ a window of $M=20$ is sufficient to capture the linearly structured information it carries.

\begin{table}[t]
\centering
\small
\begin{tabular}{lcc}
\toprule
Configuration & $M$ & GSM8K \\
\midrule
routing-only SFT        & --  & 51.99\spm{0.76} \\
\midrule
AAR$^T$               & 10  & 51.99\spm{1.13} \\
                        & 20  & \textbf{54.10}\spm{0.52} \\
                        & 128 & 53.74\spm{0.91} \\
                        & 512 & 52.96\spm{0.76} \\
\bottomrule
\end{tabular}
\caption{Window size ablation for AAR$^T$, applied in all layers. Best in bold.}
\label{tab:window}
\end{table}
\section{From Routing to Generation: Internal and External Effects of AAR}
\label{sec:analysis}

We analyze the effects of AAR on model behavior at two levels: internally, by examining how routing changes propagate to the attention mechanism, and externally, by examining how they manifest in the model's generation. For clarity, both analyses use AAR$^{\text{T}}$, which isolates the effect of the time-domain attention signal alone.

\subsection{Internal: AAR Reshapes Attention}
\label{sec:sinks}

To understand the low-level changes underlying the reasoning improvements, we examine whether and how the attention weights themselves differ between AAR$^{\text{T}}$ and baseline models. We collect attention weights generated when processing GSM8K prompts, using the AAR$^{\text{T}}$ models of Table~\ref{tab:main} with $M=20$ and their corresponding baselines.

\begin{figure}[h]
    \centering
    \includegraphics[width=\linewidth]{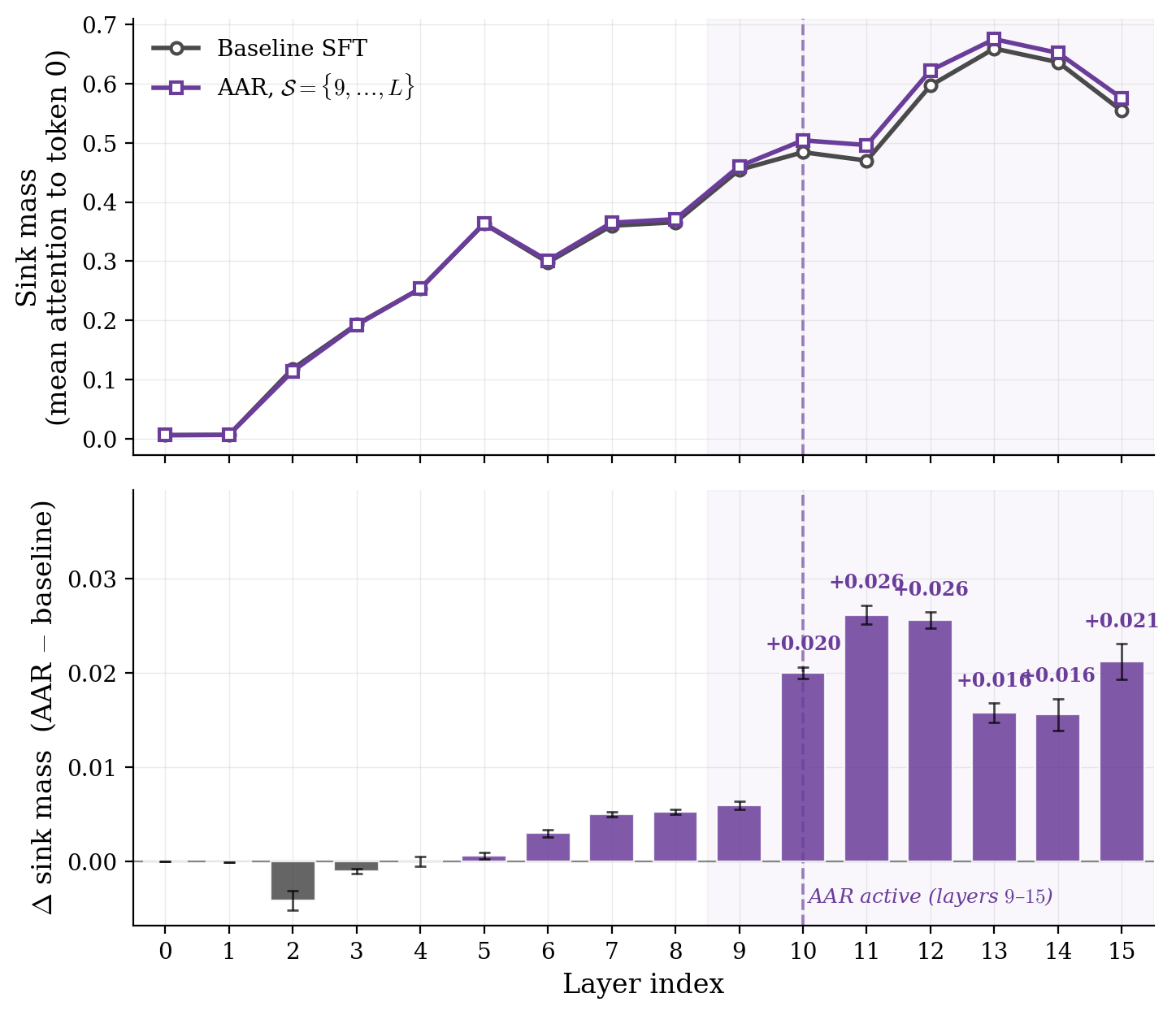}
    \caption{Top: Absolute weight to the attention sink, per layer (averaged across prompt's tokens). Bottom: the absolute difference, isolated. The different routing decisions of layer 9, because of AAR, lead to different attention weights produced in the attention submodule, at layer 10. }
    \label{fig:attention_sinks}
\end{figure}

We find that AAR$^{\text{T}}$ amplifies attention sinks, \textit{i.e.} the concentration of attention mass on the first token of the sequence \citep{xiao2024efficient}, and verify that this effect is associated with AAR via the layer-selective variant: when AAR is applied from layer $l$ onwards, attention sinks are amplified starting precisely in layer $l+1$ (Figure~\ref{fig:attention_sinks}). Routing changes at layer $l$ alter expert selection and weighting, which in turn reshapes the representations passed to layer $l+1$'s attention submodule, \textbf{without any direct update to the attention mechanism itself}.

We hypothesize that this mechanism could be one of the ways AAR improves mathematical reasoning. Prior work has established that attention sink formation stabilizes decoding by anchoring computation to a fixed reference token \citep{xiao2024efficient, liu2026sinktrack}. Under AAR, the model reinforces the pattern of concentrating attention on the first token, through routing alone. Routing thus acts not only as a selector of expert capabilities, but as an indirect regulator of attention's processing downstream.

\subsection{External: AAR Reduces Long Diverging Generation}
\label{sec:length}

We compare answer lengths on GSM8K between matched pairs of AAR$^{\text{T}}$ and baseline models in Table~\ref{tab:length}. When both models answer incorrectly, AAR-T produces substantially shorter answers (mean $-11.8\%$, P99 $-14.9\%$). When both models are correct, lengths are indistinguishable. The effect is concentrated in the wrong-answer regime and consistent across the full output length distribution, suggesting that AAR reduces diverging generation length \citep{li2026sample}: the model commits to an answer sooner when wrong, rather than generating long rambling reasoning chains. Correct answers are mostly unaffected, indicating the model has not simply learned to be uniformly brief. Answer lengths also become shorter for the answers that get corrected under AAR$^{\text{T}}$.

\begin{table}[h]
\centering
\small
\resizebox{\columnwidth}{!}{%
\begin{tabular}{llccc}
\toprule
Condition & Metric & Base & AAR-T & Rel.\ $\Delta$ \\
\midrule
\multirow{3}{*}{Both Correct}
 & Mean   & 68.2  & 67.6  & $-0.8\%$  \\
 & Median & 54.0  & 54.0  & $0.0\%$   \\
 & P99 (chars)    & 251.0 & 248.0 & $-1.2\%$  \\
\midrule
\multirow{4}{*}{Both Wrong}
 & Mean        & 163.7  & 144.4  & $-11.8\%$ \\
 & Median      & 78.0   & 73.0   & $-6.4\%$  \\
 & P99 (chars)         & 8513.0 & 7242.0 & $-14.9\%$ \\
 & Ramble Rate & 5.63\% & 4.29\% & $-23.9\%$ \\
 \midrule
 \multirow{4}{*}{Wrong $\rightarrow$ Correct}
 & Mean        & 140.4  & 89.5  & $-36.25\%$ \\
 & Median      & 83.0   & 80.0   & $-3.6\%$  \\
 & P99 (chars)         & 7306.0 & 1036.0 & $-85.8\%$ \\
\bottomrule
\end{tabular}}
\caption{Length analysis (tokens) on GSM8K averaged across seeds and context sizes $M$. Ramble Rate denotes responses exceeding 1,000 characters. AAR$^{\text{T}}$ leaves correct responses nearly unchanged while reducing the length of incorrect reasoning, especially in the tail (P99). The answer length reduction also persists for answers that get corrected by AAR$^{\text{T}}$.}
\label{tab:length}
\end{table}
\section{AAR and Layer Depth}
\label{sec:probe}
Applying AAR to every layer is not always beneficial. Under our router-only training regime, we find that doing so can substantially degrade performance on retrieval-heavy tasks, even while improving mathematical performance.
AAR's effectiveness therefore depends critically on where in the network it is applied. Across models, our results suggest that \textit{targeting middle-to-deep layers provides a robust starting point for improving mathematical reasoning}, including when transferring AAR to previously untested architectures.
This depth sensitivity is itself informative. Since AAR modifies only routing while the rest of the model remains frozen, varying its starting layer provides a controlled intervention on the routing--attention circuit at different depths. We use this intervention as a depth-wise probe and uncover a retrieval--reasoning tension: early layers are particularly sensitive for factual retrieval, whereas mathematical reasoning benefits are preserved when AAR is introduced deeper in the network.

\subsection{A Retrieval-Reasoning Tension Across Depth}

When AAR$^{\text{T}}$ is applied to all layers, MMLU performance degrades by $-2.1$ pp on average. This degradation is not uniform across subjects: Table~\ref{tab:mmlu_subjects} shows that performance on reasoning-oriented subjects (Math, Computer Science) is preserved or improved, while retrieval-heavy subjects (Biology, History, Philosophy) suffer consistent deterioration. This dissociation suggests that AAR's reasoning-amplifying behavior interferes with factual retrieval when applied in early layers.

\begin{figure}[h]
    \centering
    \includegraphics[width=\linewidth]{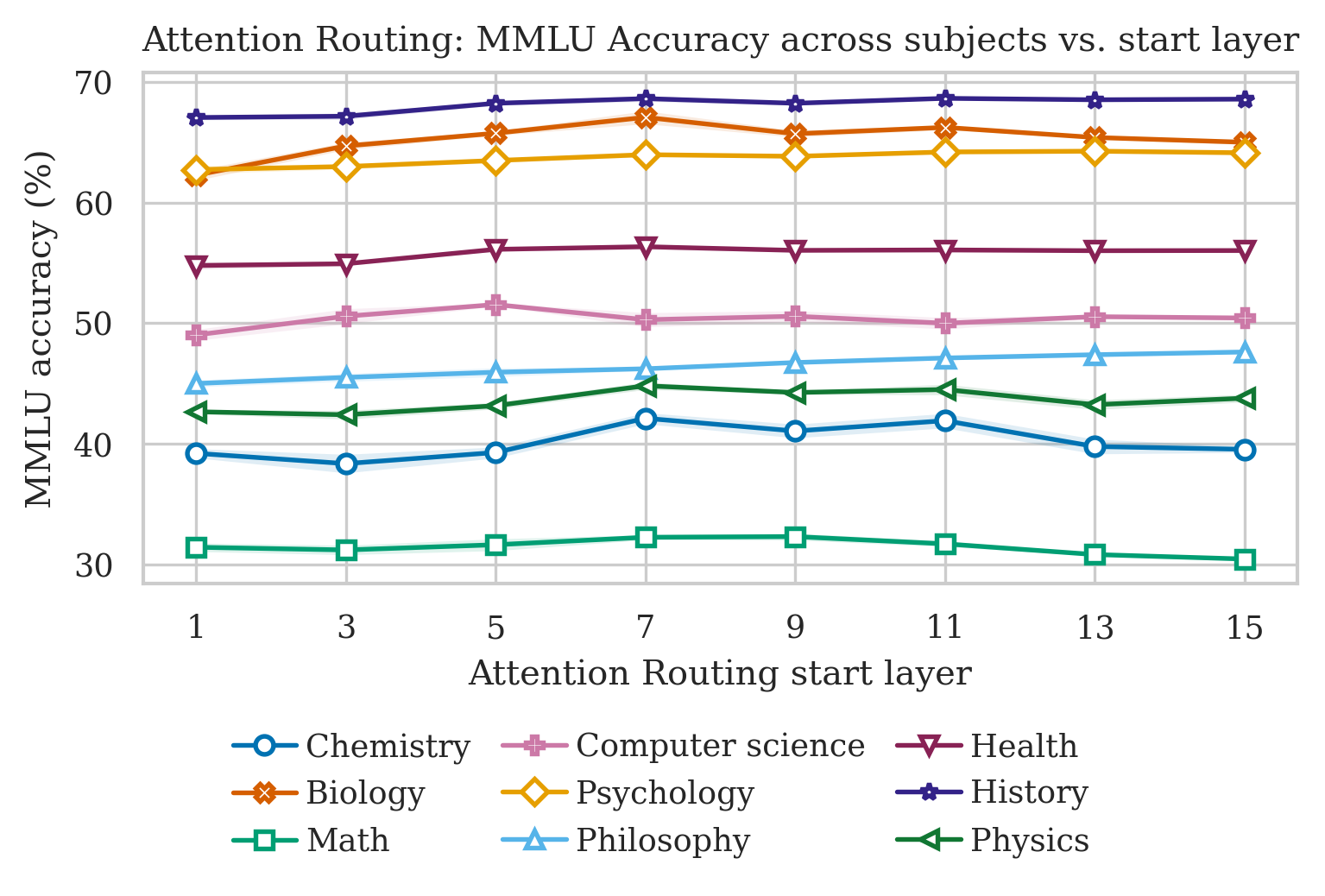}
    \caption{AAR$^{\text{T}}$ ($M=20$) accuracy across MMLU subjects as a function of starting layer. Most subjects' accuracy curves follow a bell shape.}
    \label{fig:mmlu_subjects}
\end{figure}

To localize this effect, we vary the starting layer from which AAR$^{\text{T}}$ is applied onwards and evaluate per-subject MMLU performance (Figure~\ref{fig:mmlu_subjects}). For retrieval-heavy subjects, performance is worst when AAR starts early and recovers as the starting layer increases. For reasoning-heavy subjects, improvements persist regardless of starting depth. For subjects combining both, such as Economics or Physics, performance peaks when AAR starts around layers $7$--$11$. We identify this region as the main locus of the retrieval-reasoning tension under AAR: early layers are critical for factual lookup and are sensitive to AAR-driven changes; beyond this region, retrieval-relevant structure has consolidated and AAR can be applied without disrupting it.

Per-layer attention weights thus act as a fingerprint of each layer's information processing: AAR reveals where this tension is concentrated, and that it is readable from the attention signal. This is consistent with and extends \citet{song2026demystifyingrolesllmlayers}, who show that shallow layers dominate retrieval tasks while mid-to-deep layers are essential for reasoning. Rather than probing representations post-hoc, layer-selective AAR probes the functional role of attention weights by measuring the downstream effect of routing changes directly.

\begin{table}[t]
\centering
\small
\resizebox{\columnwidth}{!}{%
\begin{tabular}{lrrr}
\toprule
MMLU Subject & $\Delta$ $M$=10 & $\Delta$ $M$=20 & $\Delta$ $M$=64 \\
\midrule
Math          & $+0.7$ & $+0.6$ & $+0.5$ \\
Comp.\ Sci.   & $+0.5$ & $+0.4$ & $-0.1$ \\
\midrule
Biology       & $-4.5$ & $-3.4$ & $-3.8$ \\
History       & $-2.5$ & $-2.4$ & $-2.2$ \\
Health        & $-1.4$ & $-1.8$ & $-1.9$ \\
Psychology    & $-1.4$ & $-1.7$ & $-1.2$ \\
Geography     & $-1.0$ & $+0.2$ & $+0.0$ \\
Philosophy    & $-3.2$ & $-3.1$ & $-3.4$ \\
\bottomrule
\end{tabular}}
\caption{MMLU subcategory $\Delta$ (absolute \% accuracy difference from baseline, 5 seeds) when AAR$^{\text{T}}$ is applied to all layers of OLMoE. Reasoning-oriented subjects (top) are improved; degradation is concentrated in retrieval-heavy subjects (bottom).}
\label{tab:mmlu_subjects}
\end{table}
\section{Discussion}
\label{sec:discussion}

\paragraph{Self-regulation through routing.}
By providing the router with a fingerprint of the model's contextual processing, as encoded in the attention weights, we enable a form of self-regulation: the routers can settle into configurations under which downstream attention patterns shift in ways that co-occur with improved task performance, making this coupling directly available during optimization. The effect is reflected internally in the reshaping of attention, and externally in the reduced length of wrong answers. It is not an explicitly supervised signal; it emerges from training with standard next-token prediction on the routing parameters alone.


\paragraph{From routing to generation.}
Our results trace a cascade: routing at layer $l$ reshapes the representations passed to layer $l+1$, which alters that layer's attention weights (evidenced by the one-layer-delayed sink amplification), and is accompanied by shorter generation on incorrect responses. That a change confined entirely to routing parameters can produce measurable effects at the level of generation behavior suggests that routing, attention, and generation are tightly coupled, and that routing is a consequential component of how the model processes information, beyond its role as a dispatching mechanism.

\section{Conclusion}
\label{sec:conclusion}






Standard MoE routers select experts based on each token's hidden state alone. We illustrate that disentangling the contextual information by augmenting the router with a sliding window of attention weights can meaningfully reshape model behavior through routing alone, improving performance on mathematics.
These routing changes propagate through the network: they amplify attention sinks one layer downstream, and at generation time, they reduce output length on incorrect answers while leaving correct ones unchanged.
The retrieval-reasoning tension we observe demonstrates that the attention-routing circuit carries qualitatively different information at different depths, with layer-selective AAR serving as a natural probe into this structure.
Taken together, our results reinforce the view that routing is not merely a dispatching mechanism, but a consequential degree of freedom: shifting routing from a local, per-token decision to one informed by the model's own attention patterns yields better mathematical reasoning performance.
\section*{Limitations}

AAR's routing mechanism requires explicit attention weight materialization, precluding the use of FlashAttention \citep{dao2022flashattention} at AAR-active layers. The eager attention fallback introduces overhead linear in sequence length and the number of AAR-active layers.

Our experiments focus on post-SFT adaptation: we train the auxiliary routers jointly with already-trained routing parameters, a setting that makes the optimal choice of AAR-active layers dependent on the model's prior training and may affect cross-model generalization. Empirically, the best layer ranges for OLMoE and Qwen both concentrate in middle-to-deep layers, but the specific subsets may differ. We hypothesize that joint training from scratch would mitigate these differences, and leave adaptation during full-parameter pretraining and SFT to future work.

\section*{Acknowledgments}
We acknowledge the EuroHPC Joint Undertaking for awarding this project access to the EuroHPC supercomputer LEONARDO, hosted by CINECA (Italy) and the LEONARDO consortium through an EuroHPC Development Access call.

\bibliography{custom}

@inproceedings{shazeer2017outrageously,
  title={Outrageously Large Neural Networks: The Sparsely-Gated Mixture-of-Experts Layer},
  author={Shazeer, Noam and Mirhoseini, Azalia and Maziarz, Krzysztof and Davis, Andy and Le, Quoc V and Hinton, Geoffrey E and Dean, Jeff},
  booktitle={ICLR},
  year={2017}
}

@article{fedus2022switch,
  title={Switch Transformers: Scaling to Trillion Parameter Models with Simple and Efficient Sparsity},
  author={Fedus, William and Zoph, Barret and Shazeer, Noam},
  journal={JMLR},
  year={2022}
}

@inproceedings{
zhou2022mixtureofexpertsexpertchoicerouting,
title={Mixture-of-Experts with Expert Choice Routing},
author={Yanqi Zhou and Tao Lei and Hanxiao Liu and Nan Du and Yanping Huang and Vincent Y Zhao and Andrew M. Dai and Zhifeng Chen and Quoc V Le and James Laudon},
booktitle={Advances in Neural Information Processing Systems},
editor={Alice H. Oh and Alekh Agarwal and Danielle Belgrave and Kyunghyun Cho},
year={2022},
url={https://openreview.net/forum?id=jdJo1HIVinI}
}

@misc{song2026demystifyingrolesllmlayers,
      title={Demystifying the Roles of LLM Layers in Retrieval, Knowledge, and Reasoning}, 
      author={Xinyuan Song and Keyu Wang and PengXiang Li and Lu Yin and Shiwei Liu},
      year={2026},
      eprint={2510.02091},
      archivePrefix={arXiv},
      primaryClass={cs.AI},
      url={https://arxiv.org/abs/2510.02091}, 
}

@inproceedings{yang-etal-2025-unveiling,
    title = "Unveiling Internal Reasoning Modes in {LLM}s: A Deep Dive into Latent Reasoning vs. Factual Shortcuts with Attribute Rate Ratio",
    author = "Yang, Yiran  and
      Sun, Haifeng  and
      Wang, Jingyu  and
      Qi, Qi  and
      Zhuang, Zirui  and
      Wang, Huazheng  and
      Ren, Pengfei  and
      Wang, Jing  and
      Liao, Jianxin",
    editor = "Christodoulopoulos, Christos  and
      Chakraborty, Tanmoy  and
      Rose, Carolyn  and
      Peng, Violet",
    booktitle = "Proceedings of the 2025 Conference on Empirical Methods in Natural Language Processing",
    month = nov,
    year = "2025",
    address = "Suzhou, China",
    publisher = "Association for Computational Linguistics",
    url = "https://aclanthology.org/2025.emnlp-main.111/",
    doi = "10.18653/v1/2025.emnlp-main.111",
    pages = "2186--2206",
    ISBN = "979-8-89176-332-6"
}

@inproceedings{nguyen-etal-2026-improving,
    title = "Improving Chain-of-Thought for Logical Reasoning via Attention-Aware Intervention",
    author = "Nguyen, Phuong Minh  and
      Huu-Tien, Dang  and
      Inoue, Naoya",
    editor = "Demberg, Vera  and
      Inui, Kentaro  and
      Marquez, Llu{\'i}s",
    booktitle = "Findings of the {A}ssociation for {C}omputational {L}inguistics: {EACL} 2026",
    month = mar,
    year = "2026",
    address = "Rabat, Morocco",
    publisher = "Association for Computational Linguistics",
    url = "https://aclanthology.org/2026.findings-eacl.152/",
    doi = "10.18653/v1/2026.findings-eacl.152",
    pages = "2917--2941",
    ISBN = "979-8-89176-386-9"
}

@inproceedings{
li2026sample,
title={Sample Smart, Not Hard: Correctness-First Decoding for Better Reasoning in {LLM}s},
author={Xueyan Li and Guinan Su and Mrinmaya Sachan and Jonas Geiping},
booktitle={The Fourteenth International Conference on Learning Representations},
year={2026},
url={https://openreview.net/forum?id=pNwCWIBHBC}
}

@inproceedings{
muennighoff2024olmoe,
title={{OLM}oE: Open Mixture-of-Experts Language Models},
author={Niklas Muennighoff and Luca Soldaini and Dirk Groeneveld and Kyle Lo and Jacob Morrison and Sewon Min and Weijia Shi and Evan Pete Walsh and Oyvind Tafjord and Nathan Lambert and Yuling Gu and Shane Arora and Akshita Bhagia and Dustin Schwenk and David Wadden and Alexander Wettig and Binyuan Hui and Tim Dettmers and Douwe Kiela and Ali Farhadi and Noah A. Smith and Pang Wei Koh and Amanpreet Singh and Hannaneh Hajishirzi},
booktitle={The Thirteenth International Conference on Learning Representations},
year={2025},
url={https://openreview.net/forum?id=xXTkbTBmqq}
}

@article{cobbe2021gsm8k,
  title={Training Verifiers to Solve Math Word Problems},
  author={Cobbe, Karl and Kosaraju, Vineet and Bavarian, Mohammad and Chen, Mark and Jun, Heewoo and Kaiser, Lukasz and Plappert, Matthias and Tworek, Jerry and Hilton, Jacob and Nakano, Reiichiro and others},
  journal={arXiv preprint},
  year={2021}
}

@inproceedings{suzgun2022bbh,
    title = "Challenging {BIG}-Bench Tasks and Whether Chain-of-Thought Can Solve Them",
    author = {Suzgun, Mirac  and
      Scales, Nathan  and
      Sch{\"a}rli, Nathanael  and
      Gehrmann, Sebastian  and
      Tay, Yi  and
      Chung, Hyung Won  and
      Chowdhery, Aakanksha  and
      Le, Quoc  and
      Chi, Ed  and
      Zhou, Denny  and
      Wei, Jason},
    editor = "Rogers, Anna  and
      Boyd-Graber, Jordan  and
      Okazaki, Naoaki",
    booktitle = "Findings of the Association for Computational Linguistics: ACL 2023",
    month = jul,
    year = "2023",
    address = "Toronto, Canada",
    publisher = "Association for Computational Linguistics",
    url = "https://aclanthology.org/2023.findings-acl.824/",
    doi = "10.18653/v1/2023.findings-acl.824",
    pages = "13003--13051"
}

@article{chen2021humaneval,
  title={Evaluating Large Language Models Trained on Code},
  author={Chen, Mark and Tworek, Jerry and Jun, Heewoo and Yuan, Qiming and de Oliveira Pinto, Henrique Ponde and Kaplan, Jared and Edwards, Harri and Burda, Yuri and Joseph, Nicholas and Brockman, Greg and others},
  journal={arXiv preprint},
  year={2021}
}

@article{zhou2023ifeval,
  title={Instruction-Following Evaluation for Large Language Models},
  author={Zhou, Jeffrey and Lu, Tianjian and Mishra, Swaroop and Brahma, Siddhartha and Basu, Sujoy and Luan, Yi and Zhou, Denny and Hou, Le},
  journal={arXiv preprint},
  year={2023}
}

@inproceedings{
xiao2024efficient,
title={Efficient Streaming Language Models with Attention Sinks},
author={Guangxuan Xiao and Yuandong Tian and Beidi Chen and Song Han and Mike Lewis},
booktitle={The Twelfth International Conference on Learning Representations},
year={2024},
url={https://openreview.net/forum?id=NG7sS51zVF}
}

@inproceedings{
liu2026sinktrack,
title={SinkTrack: Attention Sink based Context Anchoring for Large Language Models},
author={Xu Liu and Guikun Chen and Wenguan Wang},
booktitle={The Fourteenth International Conference on Learning Representations},
year={2026},
url={https://openreview.net/forum?id=Gg1aPETCL6}
}

@inproceedings{
lambert2025tulu,
title={Tulu 3: Pushing Frontiers in Open Language Model Post-Training},
author={Nathan Lambert and Jacob Morrison and Valentina Pyatkin and Shengyi Huang and Hamish Ivison and Faeze Brahman and Lester James Validad Miranda and Alisa Liu and Nouha Dziri and Xinxi Lyu and Yuling Gu and Saumya Malik and Victoria Graf and Jena D. Hwang and Jiangjiang Yang and Ronan Le Bras and Oyvind Tafjord and Christopher Wilhelm and Luca Soldaini and Noah A. Smith and Yizhong Wang and Pradeep Dasigi and Hannaneh Hajishirzi},
booktitle={Second Conference on Language Modeling},
year={2025},
url={https://openreview.net/forum?id=i1uGbfHHpH}
}

@inproceedings{
qiu2025layerwise,
title={Layerwise Recurrent Router for  Mixture-of-Experts},
author={Zihan Qiu and Zeyu Huang and Shuang Cheng and Yizhi Zhou and Zili Wang and Ivan Titov and Jie Fu},
booktitle={The Thirteenth International Conference on Learning Representations},
year={2025},
url={https://openreview.net/forum?id=eWNEqdH0vk}
}

@article{gsm8k,
  author       = {Karl Cobbe and
                  Vineet Kosaraju and
                  Mohammad Bavarian and
                  Mark Chen and
                  Heewoo Jun and
                  Lukasz Kaiser and
                  Matthias Plappert and
                  Jerry Tworek and
                  Jacob Hilton and
                  Reiichiro Nakano and
                  Christopher Hesse and
                  John Schulman},
  title        = {Training Verifiers to Solve Math Word Problems},
  journal      = {CoRR},
  volume       = {abs/2110.14168},
  year         = {2021},
  url          = {https://arxiv.org/abs/2110.14168},
  eprinttype   = {arXiv},
  eprint       = {2110.14168},
  bibsource    = {dblp computer science bibliography, https://dblp.org}
}

@inproceedings{bbh,
  author={Mirac Suzgun and Nathan Scales and Nathanael Schärli and Sebastian Gehrmann and Yi Tay and Hyung Won Chung and Aakanksha Chowdhery and Quoc V. Le and Ed H. Chi and Denny Zhou and Jason Wei},
  title={Challenging BIG-Bench Tasks and Whether Chain-of-Thought Can Solve Them},
  year={2023},
  cdate={1672531200000},
  pages={13003-13051},
  url={https://doi.org/10.18653/v1/2023.findings-acl.824},
  booktitle={ACL (Findings)},
  crossref={conf/acl/2023f}
}

@inproceedings{
hendrycks2021measuring,
title={Measuring Massive Multitask Language Understanding},
author={Dan Hendrycks and Collin Burns and Steven Basart and Andy Zou and Mantas Mazeika and Dawn Song and Jacob Steinhardt},
booktitle={International Conference on Learning Representations},
year={2021},
url={https://openreview.net/forum?id=d7KBjmI3GmQ}
}

@inproceedings{do-etal-2023-hyperrouter,
    title = "{H}yper{R}outer: Towards Efficient Training and Inference of Sparse Mixture of Experts",
    author = "Do, Truong Giang  and
      Khiem, Le  and
      Pham, Quang  and
      Nguyen, TrungTin  and
      Doan, Thanh-Nam  and
      Nguyen, Binh  and
      Liu, Chenghao  and
      Ramasamy, Savitha  and
      Li, Xiaoli  and
      Hoi, Steven",
    editor = "Bouamor, Houda  and
      Pino, Juan  and
      Bali, Kalika",
    booktitle = "Proceedings of the 2023 Conference on Empirical Methods in Natural Language Processing",
    month = dec,
    year = "2023",
    address = "Singapore",
    publisher = "Association for Computational Linguistics",
    url = "https://aclanthology.org/2023.emnlp-main.351/",
    doi = "10.18653/v1/2023.emnlp-main.351",
    pages = "5754--5765"
}

@misc{su2025underthinkingoverthinkingempiricalstudy,
      title={Between Underthinking and Overthinking: An Empirical Study of Reasoning Length and correctness in LLMs}, 
      author={Jinyan Su and Jennifer Healey and Preslav Nakov and Claire Cardie},
      year={2025},
      eprint={2505.00127},
      archivePrefix={arXiv},
      primaryClass={cs.CL},
      url={https://arxiv.org/abs/2505.00127}, 
}

@inproceedings{
lepikhin2021gshard,
title={{\{}GS{\}}hard: Scaling Giant Models with Conditional Computation and Automatic Sharding},
author={Dmitry Lepikhin and HyoukJoong Lee and Yuanzhong Xu and Dehao Chen and Orhan Firat and Yanping Huang and Maxim Krikun and Noam Shazeer and Zhifeng Chen},
booktitle={International Conference on Learning Representations},
year={2021},
url={https://openreview.net/forum?id=qrwe7XHTmYb}
}

@inproceedings{dikkala-etal-2023-benefits,
    title = "On the Benefits of Learning to Route in Mixture-of-Experts Models",
    author = "Dikkala, Nishanth  and
      Ghosh, Nikhil  and
      Meka, Raghu  and
      Panigrahy, Rina  and
      Vyas, Nikhil  and
      Wang, Xin",
    editor = "Bouamor, Houda  and
      Pino, Juan  and
      Bali, Kalika",
    booktitle = "Proceedings of the 2023 Conference on Empirical Methods in Natural Language Processing",
    month = dec,
    year = "2023",
    address = "Singapore",
    publisher = "Association for Computational Linguistics",
    url = "https://aclanthology.org/2023.emnlp-main.583/",
    doi = "10.18653/v1/2023.emnlp-main.583",
    pages = "9376--9396"
}

@inproceedings{
barbero2025why,
title={Why do {LLM}s attend to the first token?},
author={Federico Barbero and Alvaro Arroyo and Xiangming Gu and Christos Perivolaropoulos and Petar Veli{\v{c}}kovi{\'c} and Razvan Pascanu and Michael M. Bronstein},
booktitle={Second Conference on Language Modeling},
year={2025},
url={https://openreview.net/forum?id=tu4dFUsW5z}
}

@misc{qwen_moe,
    title = {Qwen1.5-MoE: Matching 7B Model Performance with 1/3 Activated Parameters"},
    url = {https://qwenlm.github.io/blog/qwen-moe/},
    author = {Qwen Team},
    month = {February},
    year = {2024}
}

@article{decoupling, title={Decoupling Knowledge and Reasoning in LLMs: An Exploration Using Cognitive Dual-System Theory}, volume={40}, url={https://ojs.aaai.org/index.php/AAAI/article/view/40723}, DOI={10.1609/aaai.v40i40.40723}, abstractNote={While large language models (LLMs) leverage both knowledge and reasoning during inference, the capacity to distinguish between them plays a pivotal role in model analysis, interpretability, and development. Inspired by dual-system cognitive theory, we propose a cognition attribution framework to decouple the contribution of knowledge and reasoning. In particular, the cognition of LLMs is decomposed into two distinct yet complementary phases: knowledge retrieval (Phase 1) and reasoning adjustment (Phase 2). To separate these phases, LLMs are prompted to generate answers under two different cognitive modes, fast thinking and slow thinking, respectively. The performance under different cognitive modes is analyzed to quantify the contribution of knowledge and reasoning. This architecture is employed to 15 LLMs across 3 datasets. Results reveal: (1) reasoning adjustment is domain-specific, benefiting reasoning-intensive domains (e.g., mathematics, physics, and chemistry) and potentially imparing knowledge-intensive domains. (2) Parameter scaling improves both knowledge and reasoning, with knowledge improvements being more pronounced. Additionally, parameter scaling make LLMs reasoning significantly more prudent, while moderately more intelligent. (3) Knowledge primarily resides in lower network layers, while reasoning operates in higher layers. Our framework not only helps understand LLMs from a &amp;quot;decoupling&amp;quot; perspective, but also provides new insights into existing research, including scaling laws, hierarchical knowledge editing, and limitations of small-scale-LLM reasoning.}, number={40}, journal={Proceedings of the AAAI Conference on Artificial Intelligence}, author={Yang, Mutian and Gao, Jiandong and Wu, Ji}, year={2026}, month={Mar.}, pages={34268–34276} }

@inproceedings{
dao2022flashattention,
title={FlashAttention: Fast and Memory-Efficient Exact Attention with {IO}-Awareness},
author={Tri Dao and Daniel Y Fu and Stefano Ermon and Atri Rudra and Christopher Re},
booktitle={Advances in Neural Information Processing Systems},
editor={Alice H. Oh and Alekh Agarwal and Danielle Belgrave and Kyunghyun Cho},
year={2022},
url={https://openreview.net/forum?id=H4DqfPSibmx}
}

@inproceedings{
loshchilov2018decoupled,
title={Decoupled Weight Decay Regularization},
author={Ilya Loshchilov and Frank Hutter},
booktitle={International Conference on Learning Representations},
year={2019},
url={https://openreview.net/forum?id=Bkg6RiCqY7},
}

@misc{qwen36_35b_a3b,
    title = {{Qwen3.6-35B-A3B}: Agentic Coding Power, Now Open to All},
    url = {https://qwen.ai/blog?id=qwen3.6-35b-a3b},
    author = {{Qwen Team}},
    month = {April},
    year = {2026}
}

@article{lightman2023lets,
      title={Let's Verify Step by Step}, 
      author={Lightman, Hunter and Kosaraju, Vineet and Burda, Yura and Edwards, Harri and Baker, Bowen and Lee, Teddy and Leike, Jan and Schulman, John and Sutskever, Ilya and Cobbe, Karl},
      journal={arXiv preprint arXiv:2305.20050},
      year={2023}
}


\end{document}